\pdfoutput=1
\documentclass[11pt]{article}
\usepackage[preprint]{acl}
\usepackage{times}
\usepackage{latexsym}
\usepackage[T1]{fontenc}
\usepackage[utf8]{inputenc}
\usepackage{microtype}
\usepackage{amsmath,amssymb}
\usepackage{booktabs}
\usepackage{graphicx}
\usepackage{multirow}
\usepackage{placeins}
\usepackage{tikz}
\usepackage{pifont}
\definecolor{evblue}{HTML}{2A78D6}\definecolor{bkvorange}{HTML}{EB6834}
\graphicspath{{figs/}}
\newcommand{\pp}{\,pp}
\newcommand{\figfont}{\fontsize{6}{7}\selectfont}

\title{Spend Bytes on Breadth: Precision--Count Trade-offs for\\Decode-Time KV Compression in Long Chain-of-Thought Reasoning}
\author{Runguo Li \\
  University of Illinois Urbana-Champaign \\
  \texttt{runguol2@illinois.edu}}

\begin{document}
\maketitle

\begin{abstract}
Reasoning models write most of their KV cache while decoding long chains of thought (CoT), so the cache has to be compressed online under a fixed memory budget. Decode-time methods mostly decide \emph{which} tokens to evict. We ask how a fixed byte budget should be split between the number of cached tokens and their precision. On three reasoning models and four math and science benchmarks, keeping more tokens at low precision beats eviction alone in 17 of 18 settings, with shorter outputs and less KV memory held over the run (memory-time). On two of the three models, most of what eviction loses comes from derailed runs, which keep reasoning until the length cap without reaching an answer; on Qwen3-8B at our tightest budget, eviction sends 91\% of AIME samples to the cap and the low-bit hybrid 40\%. The best bit-width depends on the model and budget, and the offline attention error we measured does not predict it reliably, so BreadthKV picks it with a 60-problem end-to-end calibration. Under the same protocol, a joint rate--distortion allocator (RDKV) is 1.0 point less accurate (95\% CI $-2.1$ to $+0.1$) while using 27\% more memory-time, and our re-implementation of ThinKV is about 12 points less accurate.
\end{abstract}

\section{Introduction}
\begin{figure*}[t]
\centering
\begin{tikzpicture}[font=\scriptsize, >=stealth, x=1cm, y=1cm]
\fill[black!4, rounded corners=4pt] (0,0) rectangle (7.55,4.8);
\node[anchor=west] at (0.1,4.55) {\textbf{(a)} Same bytes, different split (one KV head, at a compression step)};
\begin{scope}[yshift=0.35cm]
\draw[black!40, dashed] (1.2,1.3) -- (1.2,3.85); \draw[black!40, dashed] (4.4,1.3) -- (4.4,3.85);
\node[black!65] at (2.8,3.85) {decode-token budget: bytes of $B$ fp16 tokens};
\node[anchor=west] at (0.1,3.5) {\textbf{\textcolor{evblue}{Eviction (SnapKV-D)}}};
\fill[black!18, rounded corners=1pt] (0.1,2.8) rectangle (1.1,3.25); \node[font=\tiny, align=center] at (0.6,3.025) {prompt\\(fp16)};
\foreach \i in {0,...,7} \fill[evblue!80] (1.22+\i*0.4,2.84) rectangle ++(0.36,0.37);
\foreach \i in {0,...,6} {\draw[black!35] (4.55+\i*0.4,2.84) rectangle ++(0.36,0.37); \node[black!45, font=\tiny] at (4.73+\i*0.4,3.025) {$\times$};}
\node[anchor=north, black!75] at (2.8,2.78) {$B$ = 512 tokens, 16-bit};
\node[anchor=north, black!55] at (5.95,2.78) {evicted};
\node[anchor=west] at (0.1,2.2) {\textbf{\textcolor{bkvorange}{BreadthKV}}};
\fill[black!18, rounded corners=1pt] (0.1,1.47) rectangle (1.1,1.95); \node[font=\tiny, align=center] at (0.6,1.71) {prompt\\(fp16)};
\foreach \i in {0,...,16} \foreach \j in {0,1,2} \fill[bkvorange!85] (1.22+\i*0.155,1.48+\j*0.16) rectangle ++(0.125,0.125);
\foreach \j in {0,1,2} \fill[black!55] (3.9,1.48+\j*0.16) rectangle ++(0.45,0.125);
\foreach \i in {0,...,1} {\draw[black!35] (4.55+\i*0.4,1.53) rectangle ++(0.36,0.37); \node[black!45, font=\tiny] at (4.73+\i*0.4,1.715) {$\times$};}
\node[anchor=north, black!75, align=center, font=\figfont] at (2.5,1.45) {$B_b$ tokens at $b$ bits (KIVI format:\\K per channel, V per token)};
\node[anchor=south, black!75, font=\figfont, inner sep=1pt] at (4.12,1.97) {fp16 window};
\node[anchor=west, black!75, align=left, font=\figfont] at (5.45,1.72) {$B=512$, $b=2$:\\$B_b=2{,}412$ tokens};
\end{scope}
\draw[black!45, rounded corners=3pt, fill=white] (0.1,0.08) rectangle (7.45,0.88);
\node[anchor=west, align=left, font=\tiny] at (0.2,0.48) {\textbf{Every 128 decode steps:}\\\ding{172} quantize full 64-token blocks outside the 128-token fp16 window;\\\ding{173} evict with SnapKV-D to $B_b = 160 + (B-160)\cdot 16/(b+0.5)$ tokens.};
\begin{scope}[xshift=7.7cm]
\fill[black!4, rounded corners=4pt] (0,0) rectangle (3.85,4.8);
\node[anchor=west] at (0.1,4.55) {\textbf{(b)} Choose $b$ end to end};
\begin{scope}[yshift=0.25cm]
\node[draw=black!45, rounded corners=2pt, fill=white, align=center, inner sep=2pt] (cal) at (1.92,3.55) {calibration set: 60 AIME\\2022--23 problems $\times$ 4};
\node[black!70, font=\figfont] at (1.92,2.92) {run byte-matched $b\in\{2,3,4\}$ hybrids};
\draw[->, black!55] (cal.south) -- (1.92,3.07);
\draw[black!40] (0.45,1.3) -- (3.4,1.3);
\fill[bkvorange!85] (0.75,1.3) rectangle ++(0.55,{43.3*0.028}); \node[font=\figfont] at (1.025,{1.3+43.3*0.028+0.1}) {43.3};
\fill[black!30] (1.65,1.3) rectangle ++(0.55,{30.4*0.028}); \node[font=\figfont] at (1.925,{1.3+30.4*0.028+0.1}) {30.4};
\fill[black!30] (2.55,1.3) rectangle ++(0.55,{20.8*0.028}); \node[font=\figfont] at (2.825,{1.3+20.8*0.028+0.1}) {20.8};
\node at (1.025,1.13) {2-bit}; \node at (1.925,1.13) {3-bit}; \node at (2.825,1.13) {4-bit};
\node[black!65, font=\figfont, align=center] at (1.92,0.72) {calibration accuracy (\%)\\Qwen3-8B, $B=512$};
\node[draw=bkvorange, very thick, rounded corners=2pt, fill=white, inner sep=2pt, font=\figfont] at (1.92,0.25) {pick $b^\star=2$ (3 runs, no training)};
\end{scope}
\end{scope}
\begin{scope}[xshift=11.7cm]
\fill[black!4, rounded corners=4pt] (0,0) rectangle (4.2,4.8);
\node[anchor=west] at (0.1,4.55) {\textbf{(c)} More breadth, less derailment};
\begin{scope}[yshift=0.25cm]
\draw[black!40] (0.35,0.95) -- (4.05,0.95);
\foreach \x/\cap/\acc/\lab in {0.45/91.0/9.4/SnapKV-D, 1.38/65.4/28.8/4-bit, 2.31/54.8/38.1/3-bit, 3.24/40.4/51.7/2-bit} {
  \fill[black!45] (\x,0.95) rectangle ++(0.32,{\cap*0.024}); \node[font=\figfont, fill=white, inner sep=0.4pt, text=black!70] at ({\x+0.16},{0.95+\cap*0.024+0.1}) {\cap};
  \fill[bkvorange!85] ({\x+0.35},0.95) rectangle ++(0.32,{\acc*0.024}); \node[font=\figfont, fill=white, inner sep=0.4pt] at ({\x+0.51},{0.95+\acc*0.024+0.1}) {\acc};
  \node[font=\figfont] at ({\x+0.335},0.8) {\lab};}
\draw[->, black!55] (0.45,0.5) -- node[below, font=\figfont, inner sep=1pt] {fewer bits per token, more tokens (same bytes)} (3.95,0.5);
\fill[black!45] (1.35,3.83) rectangle ++(0.2,0.13); \node[anchor=west, font=\tiny] at (1.57,3.9) {runs hitting the 32K cap (\%)};
\fill[bkvorange!85] (1.35,3.6) rectangle ++(0.2,0.13); \node[anchor=west, font=\tiny] at (1.57,3.67) {accuracy (\%)};
\node[anchor=east, font=\figfont, black!65] at (4.1,3.4) {Qwen3-8B, AIME24/25, $B=512$};
\end{scope}
\end{scope}
\end{tikzpicture}
\caption{Overview of BreadthKV. (a) At the same KV byte budget per head, eviction keeps $B$ full-precision decode tokens, whereas BreadthKV quantizes older tokens to $b$ bits in the KIVI format and keeps $B_b$ of them plus a 128-token fp16 window; both compress every 128 decode steps and keep the prompt in full. (b) The bit-width is chosen for each model and budget by running the byte-matched $\{2,3,4\}$-bit hybrids on 60 held-out AIME problems. (c) At equal bytes, fewer bits per token lower the share of runs that reach the 32K cap and raise accuracy (FullKV: 10.2\% capped, 69.8\% accuracy). Squares are not to scale.}\label{fig:overview}
\end{figure*}
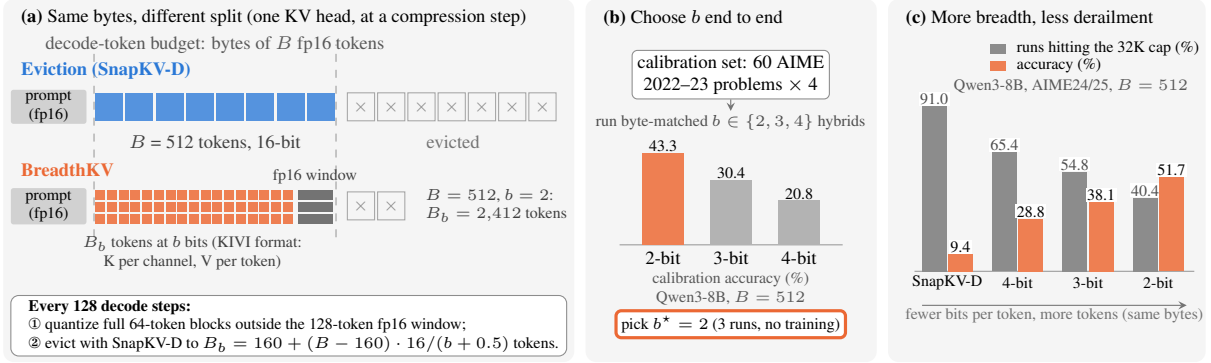

On competition math, long-CoT reasoning models often generate 10--30K tokens per problem, and the KV cache grows with every decoded token. At 32K tokens, a single Qwen3-8B sequence holds 4.8\,GB of KV cache in bf16, so four sequences already take more memory than the model's 16\,GB of weights. The cache has to be compressed online, and there is no fixed query to guide it, as there is when compressing a prompt. Token importance also shifts during decoding, since reflection steps often return to early intermediate results, and errors accumulate over the trace. Under eviction, models can fall into loops that run until the length limit~\citep{vase,kvrescue}; in one of our Qwen3-8B runs, the model finds the right answer within the first tenth of its output and then re-checks it until the 32K cap (Appendix~\ref{app:example}), and a smaller cache can make the model write more tokens, which offsets part of the memory saved~\citep{holdonto,reco}.

Recent decode-time methods decide \emph{which} tokens to keep, using attention scores~\citep{h2o,snapkv}, redundancy-aware scores~\citep{rkv}, thought-type heuristics~\citep{thinkv}, or value ranges combined with stochastic eviction~\citep{vase}. Concurrent work finds that uniform random eviction, with the prompt kept, does about as well as these scores at 4$\times$ compression~\citep{randomattn}.

A simple offline measurement shows why the other axis, precision, deserves attention. Even an oracle that knows which tokens future queries will attend to loses about 13\% of their attention mass when it keeps 10\% of the tokens (\S\ref{sec:pilot}). A long CoT keeps going back to its own intermediate results, and an evicted token cannot be recovered. A token stored at 2 bits is noisy, but a later reflection step can still find it. At a fixed byte budget, a head can keep $K$ tokens in fp16, about $3K$ at 4 bits or about $6K$ at 2 bits, so the question becomes how to trade depth (precision per token) against breadth (the number of tokens kept). \citet{moretokens} studied this trade-off for long inputs, where pruning plus quantization beats pruning at equal memory. Long-CoT decoding differs in two ways: the cache is written during generation, so errors compound over tens of thousands of steps, and a lossy cache changes what the model writes, including how long it runs. \citet{notallbits} compare quantization and eviction for reasoning models, but separately, and find 2-bit quantization lossy.

We combine the two at equal KV bytes and much tighter budgets. On three models and four benchmarks, the byte split matters more than the choice among the six eviction methods we tested. With the bit-width chosen by a short end-to-end calibration, a method we call BreadthKV, the low-bit hybrid scores above SnapKV-style eviction in 17 of 18 settings (\S\ref{sec:main}). The best bit-width depends on the model and budget. No fixed width is safe for every model, and neither the offline attention error we measured nor a per-head online probe finds the right one (\S\ref{sec:bits}--\ref{sec:calib}). On two of the three models, eviction loses mostly through derailed runs that never conclude, and keeping more tokens at lower precision makes them rarer, down to about 2 bits with the codecs we tested (\S\ref{sec:derail}--\ref{sec:rescue}). We also run RDKV~\citep{rdkv} with the authors' code, where we found and fixed a zero-point clamp in the released quantizer, and our re-implementation of ThinKV~\citep{thinkv} under the same protocol (\S\ref{sec:rdkv}--\ref{sec:thinkv}). Last, we time a low-bit decode kernel (\S\ref{sec:eff}).

\section{Related work}
\textbf{Which tokens to keep.} Most KV compression work selects tokens, and for long inputs it does so by position, accumulated attention or value-aware scores~\citep{streamingllm,h2o,snapkv,pyramidkv,pyramidinfer,curdkv}, and Ada-KV~\citep{adakv} splits the budget across heads. Reasoning models need eviction during decoding, and a fast-growing line of methods targets that case~\citep{rkv,lazyeviction,raas,gkv,triattention,epikv,lethe,reasonalloc,rlkv}. Two papers are most relevant here. Random Attention~\citep{randomattn} shows that random eviction with the prompt kept matches the scored methods at about 4$\times$ compression, and stays tied with TriAttention at tighter budgets while VaSE falls behind, which suggests that better scores are not where the remaining accuracy lies. VaSE~\citep{vase} finds that evicting a few value states with large ranges makes generations loop, restating intermediate reasoning without progress, and names a high-precision reserve for such values as future work, which hints that precision and looping are linked. All of these methods keep the selected tokens at full precision. We fix the scorer (SnapKV-D) and vary the precision. SeerAttention-R~\citep{seerattnr} learns sparse attention for long decoding but keeps the full cache, so it saves compute and not memory.

\textbf{How many bits.} KIVI~\citep{kivi} and KVQuant~\citep{kvquant} quantize keys per channel and values per token, the format we use. Later work allocates bits more finely, by layer~\citep{kvtuner}, channel~\citep{kitty}, token~\citep{dontwaste,triaxialkv,zipcache}, query-aware channel selection for reasoning~\citep{mixkvq}, or over time, as PM-KVQ~\citep{pmkvq} lowers precision while a long CoT grows, and other work improves the quantizer itself~\citep{kvarn,ratequant,sphericalkv,quarot,turboquant} or applies transform coding~\citep{kvtc}. Without eviction, even a 2-bit cache of a 32K-token trace does not fit our budgets, so these methods are complementary to our question. BreadthKV uses one bit-width per model and budget, and \S\ref{sec:rdkv} tests whether finer allocation helps. Low-rank methods~\citep{palu,xkv,starkv,shadowkv} compress the hidden dimension instead, but decode-time values in our pilot are far from low-rank (\S\ref{sec:pilot}).

\textbf{Trading precision for count (closest work).} \emph{More Tokens, Lower Precision}~\citep{moretokens} shows that at equal memory, pruning combined with quantization, for example 4$\times$ the tokens at 4 bits, outperforms pruning alone. The study covers long-input prefill (LongBench, NIAH, RULER) on non-reasoning models, where 2-bit precision degrades sharply and the headline configuration stores 4$\times$ the tokens at 4 bits. We study decode-time compression of long CoT, in which the cache is written during generation and a lossy cache changes how long the model runs. At our budgets, calibration picks 2 or 3 bits in 5 of 6 model--budget pairs, and always using 4 bits is significantly worse (\S\ref{sec:calib}). \emph{Not All Bits Are Equal}~\citep{notallbits} compares eviction (R-KV, StreamingLLM) and quantization (HQQ, per channel) separately on Qwen3 models from 0.6B to 32B. It counts weight memory in the budget and uses eviction budgets of 2K--8K tokens. It finds that eviction beats quantization for models below the effective size of an 8-bit 8B model and that 2-bit quantization stays lossy. Our 7--8B bf16 models are above that threshold, we combine quantization and eviction at equal KV bytes, and our eviction budgets (512--1024 tokens) are 2--16$\times$ smaller. Joint allocators decide per token or element between several precisions and eviction. MiKV~\citep{mikv} keeps evicted tokens at low precision, HqeKV~\citep{hqekv} optimizes the action for each element, QEvict~\citep{qevict} keeps fp16, INT2 and evicted tiers on long inputs, ThinKV~\citep{thinkv} works by thought type, and RDKV~\citep{rdkv} solves a rate--distortion allocation. ThinKV reports that hybrids trace a Pareto frontier and avoid the length inflation of pure quantization, which reaches 5.1$\times$ on R1-Llama-8B. These works propose allocators, whereas we vary only the byte split, with the scorer and format fixed, and evaluate ThinKV and RDKV under the same protocol.

\textbf{Length inflation and loops.} Compression lengthens reasoning, both for KV eviction, as Hold Onto That Thought~\citep{holdonto} and ReCo~\citep{reco} report, and for weight quantization~\citep{quantinflates}, and \citet{waitloop} study why reasoning models loop in the first place. KV-Rescue~\citep{kvrescue} separates incoherent generation from repetition loops under eviction and repairs runs with reasoning steps from a small full-context model, and SkipKV~\citep{skipkv} suppresses redundant sentences. \citet{accevidence} show that compression can keep final answers intact while degrading the reasoning behind them. We add a controlled measurement of how the share of runs that reach the cap moves with the byte split (\S\ref{sec:derail}) and test whether forcing a conclusion helps (\S\ref{sec:rescue}).

Random Attention, QEvict, ReCo, KV-Rescue and \citet{accevidence} appeared within three months of this work and are concurrent with it.

\section{Byte-budgeted precision--count compression}\label{sec:method}
\textbf{Budget.} For each sequence, layer and KV head, the keys and values of decode tokens are limited to the bytes of $B$ fp16 tokens. The prompt is kept in full and not counted, for every method. All methods compress every 128 decode steps (VaSE every 64, as in its code) and bring each head back to its budget. We also report KV memory-time, the bytes held summed over decode steps, which counts both the budget and the output length.

\textbf{Eviction (SnapKV-D).} Decode-time SnapKV keeps each head's top-$B$ decode tokens by the attention they receive from the last 8 queries, max-pooled over 7 positions, and always keeps the 8 most recent tokens. R-KV, H2O, StreamingLLM and random eviction follow the same schedule with their own scores.

\textbf{Precision-for-count hybrid ($b$-bit + SnapKV-D).} The hybrid uses the same eviction, but decode tokens older than a 128-token fp16 residual window are quantized in complete 64-token blocks in the KIVI format~\citep{kivi}. K is quantized per channel within each 64-token group and V per token within 64-channel groups. Both use asymmetric quantization with an fp16 scale and zero point per group, which costs $b+0.5$ bits per element. To match bytes, we raise the token budget to
\begin{equation}
B_b = 160 + (B-160)\cdot\frac{16}{b+0.5},
\end{equation}
where 160 approximates the fp16 tail (128--191 tokens), so hybrids stay within about 5\% of the budget; memory-time is measured exactly. Rounded down, this gives 1411/1769/2412 tokens at 4/3/2 bits at $B=512$, and 3232/4109/5689 at $B=1024$.

\textbf{BreadthKV: choosing $b$ by end-to-end calibration.} We run the byte-matched $\{2,3,4\}$-bit hybrids on a held-out calibration set (60 AIME 2022--23 problems $\times$ 4 samples) and keep the most accurate $b$ for each model and budget, breaking ties toward fewer bits. Calibration takes three runs per model and budget, needs no training and uses no test problems. We call the byte-matched hybrid with the calibrated $b$ BreadthKV and use it as the reference in all comparisons.

\section{Pilot analyses}\label{sec:pilot}
Before running anything end to end, we checked offline whether breadth can help at all. We take teacher-forced FullKV traces of R1-Distill-Qwen-7B on 8 MATH500 problems and cut each trace at 30\%, 60\% and 90\% of its length. At each cut, the KV of decode tokens older than the recent window is compressed to a fraction $r$ of its fp16 bytes, and we measure the relative error of the attention output of the next 256 queries over all layers (Table~\ref{tab:pilot}). Quantizing to 4 bits and evicting with a simple window score gives lower error than oracle eviction at every byte ratio (0.16 vs.\ 0.19 at 5\% of the bytes). Eviction also has a floor that no scorer can remove. Even with exact knowledge of future attention, it loses about 13\% of the future attention mass at 10\% retention. Low-rank projection fits these caches poorly, since V needs 78 of its 128 dimensions to keep 90\% of its energy, and its top-16 subspace drifts between the first and last thirds of a trace (overlap 0.69).

\begin{table}[t]
\centering\footnotesize\setlength{\tabcolsep}{2.5pt}
\caption{Relative attention-output error at equal bytes (R1-Distill-Qwen-7B, 8 MATH500 traces). Window: SnapKV-style window attention.}\label{tab:pilot}
\begin{tabular}{lcccccc}
\toprule
& \multicolumn{3}{c}{eviction} & & \multicolumn{2}{c}{$b$-bit + eviction} \\
\cmidrule(lr){2-4}\cmidrule(lr){6-7}
bytes & random & window & oracle & low-rank & 2-bit & 4-bit \\
\midrule
5\% & 0.403 & 0.244 & 0.191 & 0.314 & 0.206 & \textbf{0.162} \\
10\% & 0.365 & 0.200 & 0.144 & 0.284 & 0.175 & \textbf{0.104} \\
20\% & 0.298 & 0.146 & 0.093 & 0.249 & -- & \textbf{0.046} \\
\bottomrule
\end{tabular}
\end{table}

\section{Experiments}\label{sec:exp}
\textbf{Setup.} We evaluate R1-Distill-Qwen-7B, R1-Distill-Llama-8B~\citep{deepseekr1} and Qwen3-8B in thinking mode~\citep{qwen3}. We report pass@1 on AIME24 and AIME25 with 8 samples per problem, on MATH500~\citep{mathds,lightman} with one sample, and on GPQA-Diamond~\citep{gpqa} with 4 samples (R1-Qwen only). We decode with temperature 0.6 and top-$p$ 0.95 (top-$k$ 20 for Qwen3), up to 32K tokens (16K for MATH500), at budgets $B\in\{512,1024\}$. AIME24 with R1-Qwen served as our dev set while we developed the methods (Appendix~\ref{app:pact}), so we exclude it from the test results. For statistics, we use a paired bootstrap over problems (5000 resamples, 95\% CI); pooled tests over settings resample problems within each setting. We use our own decoding engine on HF weights, with FlashAttention-2 decoding~\citep{flashattn2}, CUDA graphs and per-head variable-length caches. The engine matches HF (teacher-forced top-1 agreement $\ge$98\%) and reproduces the reported FullKV accuracy of R1-Qwen (MATH500 93.0 vs.\ 92.8; AIME24 55.4 vs.\ 55.5). Appendices~\ref{app:setup} and~\ref{app:full} give experimental details and all results.

\subsection{Main results}\label{sec:main}

\begin{figure*}[t]
\centering
\includegraphics[width=\textwidth]{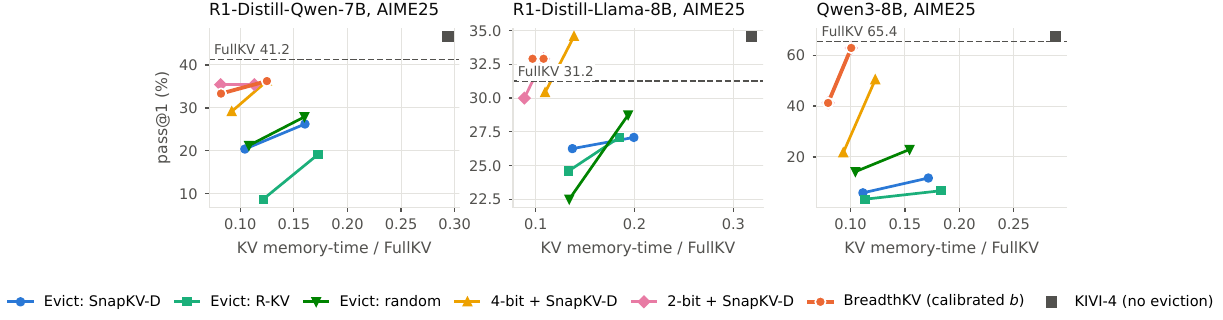}
\caption{Accuracy vs.\ KV memory-time (relative to FullKV) on AIME25. Each line connects a method's results at $B=512$ and 1024; upper left is better. BreadthKV uses the calibrated bit-width, so its points coincide with the hybrid of that width where one is plotted (on Qwen3-8B it covers the 2-bit line). The square is KIVI-4 without eviction, and the dashed line is FullKV.}\label{fig:main}
\end{figure*}
Keeping more tokens at low precision beats keeping fewer tokens in fp16, and on Qwen3-8B the difference is large. Table~\ref{tab:main} gives BreadthKV's gain over SnapKV-D at the same budget. The gain is positive in 17 of 18 settings, its CI excludes zero in 16 (15 after a Holm correction over the 18 tests), and pooled over all settings it is 17.4\pp{} [15.6, 19.3], or 8.6, 7.9 and 35.8\pp{} on R1-Qwen, R1-Llama and Qwen3. The one negative case is R1-Llama MATH500 at $B=1024$, where SnapKV-D is already within 2.4\pp{} of FullKV. Choosing the bit-width per setting on the test set, an oracle, would add only 0.83\pp{} on average (\S\ref{sec:calib}). Pure eviction nearly collapses on Qwen3-8B, which scores 5.8\% on AIME25 at $B=512$ with SnapKV-D and 65.4\% with FullKV, while BreadthKV at $B=1024$ comes close to FullKV (Appendix~\ref{app:full}).

\begin{table}[t]
\centering\small
\caption{Accuracy gain (pp) of BreadthKV over SnapKV-D at the same budget, $B=512$ / 1024. Computed from unrounded accuracies. $^*$95\% CI includes 0. $^\dagger$R1-Qwen on AIME24 is our dev set and is not reported; all other cells are test results.}\label{tab:main}
\setlength{\tabcolsep}{2.5pt}\footnotesize
\begin{tabular}{lccc}
\toprule
benchmark & R1-Qwen-7B & R1-Llama-8B & Qwen3-8B \\
\midrule
AIME24 & dev$^\dagger$ & +20.8 / +8.3 & +49.2 / +50.0 \\
AIME25 & +12.9 / +10.0 & +6.7$^*$ / +5.8 & +35.4 / +51.2 \\
MATH500 & +8.6 / +5.8 & +6.8 / $-0.8^*$ & +17.4 / +11.6 \\
GPQA & +7.3 / +6.8 & -- & -- \\
\bottomrule
\end{tabular}
\end{table}

Among the six eviction methods we tested, the choice matters much less. BreadthKV is above the best of them (SnapKV-D, R-KV, H2O, StreamingLLM, random and VaSE) in 17 of 18 settings. Random eviction matches SnapKV-D on the R1-Distill models and does better on Qwen3-8B, in line with \citet{randomattn}. VaSE~\citep{vase}, ported from its released code, is 0.6\pp{} below SnapKV-D pooled [$-1.5$, $+0.3$], and H2O is the weakest (we ran StreamingLLM and H2O at $B=1024$ only). A rank-32 low-rank baseline with a frozen basis collapses. Without eviction, KIVI-4 stays within 5.5\pp{} of FullKV, but it holds 0.29--0.34$\times$ FullKV's memory-time, against 0.08--0.22$\times$ for the byte-matched hybrids.

\textbf{Length and memory-time.} SnapKV-D outputs are 1.1--2.0$\times$ as long as FullKV's, and BreadthKV's are 0.76$\times$ as long as SnapKV-D's (pooled). The memory a cache occupies is the bytes it holds over time, so longer outputs offset part of the savings that retention ratios suggest. On Qwen3 AIME25 at $B=1024$, SnapKV-D holds 0.17$\times$ FullKV's memory-time and scores 11.7\%, while BreadthKV holds 0.10$\times$ and scores 62.9\% (Fig.~\ref{fig:main}).

\subsection{The best bit-width depends on budget and model}\label{sec:bits}
No single bit-width is best everywhere. On our dev set, the best width shared by all heads rises with the budget, from 2 bits at $B=256$ to 8 bits at $B=2048$ (Appendix~\ref{app:pact}). On the test sets the preference depends on the model. For the R1-Distill models, the 2-, 3- and 4-bit hybrids are close at $B=1024$, while Qwen3-8B clearly prefers 2 bits (Table~\ref{tab:full}). Qwen3-8B also writes the longest AIME traces (15.5--18.5K tokens on average, against 13--14K), which fits the idea that longer traces need more coverage, though three models cannot separate trace length from other differences.

\textbf{Per-head online allocation did not help.} Our first approach chose the width online for each head (PACT, Appendix~\ref{app:pact}). Offline, letting each head pick its own width lowers attention error by 20--42\%, and a cheap probe on recent queries recovers most of that gain, but end to end the per-head version scored below a fixed 4-bit width on the dev set and was indistinguishable from fixed widths on the test sets (1 of 18 comparisons significant). A width table tuned on the dev set did not transfer either; on Qwen3-8B it trails the 2-bit hybrid in all six AIME and MATH500 settings. More generally, attention error at a single step predicts accuracy poorly. Random eviction, for example, has 1.8$\times$ the attention error of window-based eviction and the same dev accuracy.

\begin{table*}[t]
\centering\footnotesize\setlength{\tabcolsep}{3pt}
\begin{minipage}{0.52\textwidth}\centering
\caption{Calibration accuracy (\%) of 2/3/4-bit hybrids and the chosen $b$.}\label{tab:calib}
\begin{tabular}{lcc}
\toprule
model & @512 & @1024 \\
\midrule
R1-Qwen-7B & 28.3/\textbf{31.2}/28.3 $\to$ 3 & 33.8/35.0/\textbf{40.0} $\to$ 4 \\
R1-Llama-8B & 33.8/\textbf{36.7}/35.8 $\to$ 3 & \textbf{40.0}/\textbf{40.0}/39.2 $\to$ 2 \\
Qwen3-8B & \textbf{43.3}/30.4/20.8 $\to$ 2 & \textbf{62.9}/57.1/50.0 $\to$ 2 \\
\bottomrule
\end{tabular}
\end{minipage}\hfill
\begin{minipage}{0.44\textwidth}\centering
\caption{Held-out regret relative to the best bit-width in each setting (18 settings). The per-setting best is picked on the test set, so all regrets are slightly overstated.}\label{tab:regret}
\begin{tabular}{lccc}
\toprule
strategy & mean & worst & $>$3\pp{} \\
\midrule
BreadthKV (calibrated) & 0.83 & 3.4 & 1/18 \\
always 2-bit & 1.21 & 7.5 & 2/18 \\
always 3-bit & 2.94 & 14.6 & 4/18 \\
always 4-bit & 5.34 & 26.2 & 7/18 \\
\bottomrule
\end{tabular}
\end{minipage}
\end{table*}

\subsection{Choosing the bit-width by end-to-end calibration}\label{sec:calib}
\begin{figure}[t]
\centering
\includegraphics[width=\linewidth]{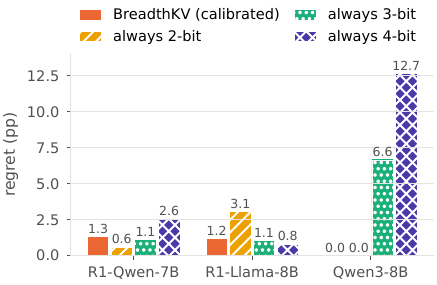}
\caption{Held-out regret of bit-width selection strategies: the mean gap (pp) to the best bit-width over each model's six test settings (two budgets $\times$ three benchmarks); lower is better.}\label{fig:regret}
\end{figure}

Since neither the offline error we measured nor online probes find the best width, we measure it directly. Offline distortion curves at equal bytes, computed on 6 FullKV AIME24 traces per model (not used to choose $b$), get the preference wrong on two of the three models; for Qwen3-8B they favor 4 over 2 bits, the opposite of what happens end to end. Three calibration runs per model and budget pick the widths in Table~\ref{tab:calib}, and Table~\ref{tab:regret} and Fig.~\ref{fig:regret} show how the choice does on the 18 held-out settings.

Calibration beats always using 4 bits by 4.5\pp{} pooled [+3.2, +5.8] and always using 3 bits by 2.1\pp{} [+1.2, +3.1]. It is not significantly better than always using 2 bits (+0.4\pp{} [$-0.2$, +1.0]). Calibration matters because each fixed width does badly on some model, 2 bits on R1-Llama (mean regret 3.1\pp{}) and 4 bits on Qwen3 (12.7\pp{}), while calibration stays within 1.3\pp{} on every model. On Qwen3, where a wrong choice hurts most, the choice is also stable; resampling the calibration problems picks 2 bits in 100\% and 99\% of resamples at the two budgets. On the R1-Distill models the choice is noisy (standard error about 5 points, clustered by problem; the chosen width is never significantly ahead of the runner-up), but the widths are also close on the test sets there, so the noise costs little. At these budgets, always using 2 bits is an equally good default, and calibration is a cheap check against the model-specific failures of any fixed width, including the 4 bits that \citet{moretokens} use for long inputs.

\subsection{Comparison with a joint rate--distortion allocator (RDKV)}\label{sec:rdkv}
RDKV~\citep{rdkv} is the prior method closest to our view of trading precision for count, so we reproduce it with the authors' code. It treats eviction and quantization as one rate--distortion problem. For each KV head, a Lagrangian knapsack assigns each V token and each K channel a width in $\{0,2,4,8,16\}$ bits, where 0 means the token is evicted or the channel is pruned, and weights a calibrated distortion table $\epsilon(b)$ by attention-based importance.

\textbf{Reproduction.} The released code compresses the prompt once after prefill and includes an experimental block-wise mode for decoding, which we use. Each 128-token block of decode tokens is scored with the queries of the next block, allocated once by the same knapsack under a per-block budget of $\beta$ fp16-token equivalents, and then frozen. We call the upstream scoring, knapsack solver and $\epsilon$ calibration directly, and run the calibration on traces from our calibration set. Because the block mode has no total budget, we choose $\beta$ for each model and budget so that RDKV's mean KV memory-time on the calibration set matches that of BreadthKV. This gives $\beta=4$ at $B=512$ and $\beta=13/15/11$ at $B=1024$ for R1-Qwen/R1-Llama/Qwen3.

\textbf{The released quantizer clamps the zero point.} The released \texttt{quantize\_and\_pack} sets the zero point to $\mathrm{clamp}(\mathrm{round}(-\min/s), 0, 2^b-1)$, which clips groups whose range does not contain 0. Post-RoPE low-frequency K channels often keep the same sign throughout a 128-token block. For such channels, even the 8-bit K error is 0.3--0.45 instead of $\approx$0.004, and generation degenerates. In the prefill setting the code was written for, a channel spans thousands of tokens and its range almost always contains 0, which probably hid the issue. Our fix stores the minimum as the offset and uses the same bits and metadata. With the released quantizer, RDKV scores 5--13\pp{} below the fixed version on AIME25 for R1-Qwen and Qwen3 and collapses on R1-Llama at both budgets (0.0\% vs.\ 31.7/35.8\%, with 73/57\% of samples capped). All other RDKV results use the fixed quantizer.

\begin{table}[t]
\centering\footnotesize\setlength{\tabcolsep}{3pt}
\caption{BreadthKV (BKV) vs RDKV (block-wise, fixed quantizer): accuracy (\%) averaged over each model's three test benchmarks (R1-Qwen: AIME25, MATH500, GPQA; others: AIME24, AIME25, MATH500), and RDKV's KV memory-time relative to BreadthKV (MT).}\label{tab:rdkv}
\begin{tabular}{llcccc}
\toprule
model & $B$ & BKV & 2-bit & RDKV & MT \\
\midrule
R1-Qwen & 512 & 51.8 & 53.8 & 49.3 & 1.20$\times$ \\
R1-Qwen & 1024 & 55.0 & 54.5 & 55.2 & 1.14$\times$ \\
R1-Llama & 512 & 58.3 & 54.6 & 54.5 & 1.18$\times$ \\
R1-Llama & 1024 & 56.4 & 56.4 & 57.9 & 1.44$\times$ \\
Qwen3 & 512 & 64.6 & 64.6 & 65.2 & 1.30$\times$ \\
Qwen3 & 1024 & 77.1 & 77.1 & 75.1 & 1.37$\times$ \\
\bottomrule
\end{tabular}
\end{table}

Per-token and per-channel allocation did not improve accuracy here, and it used more memory. Pooled over the 18 test settings, RDKV is 1.0\pp{} less accurate than BreadthKV (95\% CI [$-2.1$, $+0.1$]; a TOST test with a margin of $\pm$2\pp{}, chosen after the fact, would call them equivalent, one with $\pm$1\pp{} would not) while holding 1.27$\times$ its KV memory-time; it is ahead in 3 of the 6 model--budget averages (Table~\ref{tab:rdkv}). The per-block budget was matched on calibration-set lengths, but the cache grows with the generation and RDKV writes 6\% longer outputs, so at equal test-time memory RDKV would need a smaller block budget; the comparison as run favors RDKV.

\subsection{Comparison with a re-implementation of ThinKV}\label{sec:thinkv}
ThinKV~\citep{thinkv} uses attention sparsity to classify segments of the CoT as reasoning (R), execution (E) or transition (T) thoughts, and quantizes each type at its own precision. The R4E4T2 setting uses NVFP4 for R and E and ternary values for T, in groups of 16 with FP8 scales. At every transition, ThinKV anneals each earlier segment from 128 to 64, \ldots, 4 tokens, keeping tokens chosen by K-means on the keys. No code is available, so we re-implemented it from the paper, converting its token budget to our byte budget and keeping the prompt in full.

\textbf{Reproduction choices.} We fix the details the paper leaves open and set the type thresholds at the tertiles of ThinKV's own runtime sparsity on the calibration set, because thresholds computed offline do not transfer to decoding under compression (Appendix~\ref{app:thinkv}).

\begin{figure*}[t]
\centering
\includegraphics[width=\textwidth]{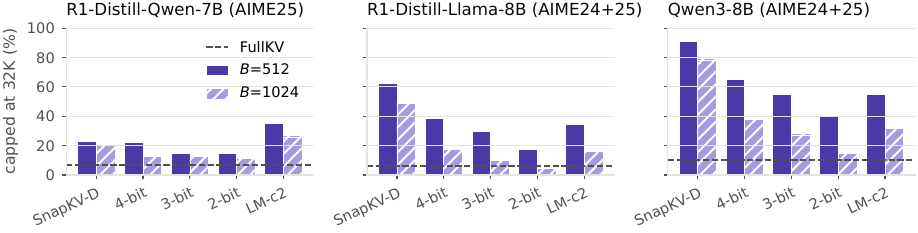}
\caption{Share of samples hitting the 32K length cap on the test AIMEs (R1-Qwen: AIME25; others: AIME24 and AIME25), by method and budget. Dashed: FullKV.}\label{fig:derail}
\end{figure*}
\textbf{Key format on R1-Distill-Qwen.} The keys of R1-Qwen carry large per-channel offsets from the k\_proj bias ($|K|$ up to $\sim$430). Quantizing the keys with any of the paper's formats (FP8 with a per-tensor scale, its 8-bit format, or symmetric NVFP4 or ternary) makes generation collapse, and AIME25 accuracy with the R4E4T2 setting is 0.0\% at both budgets. Keeping the keys in fp16 restores normal generation. Subtracting each segment's per-channel mean before quantization also fixes the collapse. The mean is stored in fp16 once per segment, but our byte count charges it as 0.125 bit per retained key element, which undercounts it once a segment is annealed; we bound this below. All our main ThinKV results use this centering.

\begin{table}[t]
\centering\footnotesize\setlength{\tabcolsep}{3pt}
\caption{BreadthKV (BKV) vs our re-implementation of ThinKV (with K centering): accuracy (\%) averaged over each model's three test benchmarks (as in Table~\ref{tab:rdkv}), ThinKV's KV memory-time relative to BreadthKV (MT; as counted, which undercounts ThinKV's per-segment K means, see text), and the share of samples hitting the length cap (\%; 32K, MATH500 16K).}\label{tab:thinkv}
\begin{tabular}{llccccc}
\toprule
& & \multicolumn{2}{c}{accuracy} & & \multicolumn{2}{c}{capped} \\
\cmidrule(lr){3-4}\cmidrule(lr){6-7}
model & $B$ & BKV & ThinKV & MT & BKV & ThinKV \\
\midrule
R1-Qwen & 512 & 51.8 & 46.0 & 0.94$\times$ & 8 & 22 \\
R1-Qwen & 1024 & 55.0 & 52.2 & 0.74$\times$ & 6 & 12 \\
R1-Llama & 512 & 58.3 & 43.9 & 0.87$\times$ & 22 & 54 \\
R1-Llama & 1024 & 56.4 & 54.3 & 0.91$\times$ & 5 & 31 \\
Qwen3 & 512 & 64.6 & 39.6 & 0.86$\times$ & 30 & 61 \\
Qwen3 & 1024 & 77.1 & 53.6 & 0.86$\times$ & 11 & 42 \\
\bottomrule
\end{tabular}
\end{table}

ThinKV scores well below BreadthKV, by 12.3\pp{} pooled [$-14.1$, $-10.5$], and the gap is largest on Qwen3, the model with the longest traces (Table~\ref{tab:thinkv}). The gap is not a matter of memory. As counted, ThinKV holds 0.86$\times$ BreadthKV's KV memory-time (range 0.72--1.01$\times$), and at most 0.93$\times$ (range 0.74--1.10$\times$) if every segment's K means are charged in full. It uses less memory because annealing at each transition shrinks old segments whatever the budget. A difference of 7--14\% is small next to what doubling the memory does, which changes BreadthKV's mean accuracy by $-2$ to $+13$\pp{} per model. Where ThinKV does differ is derailment (\S\ref{sec:derail}), since its runs reach the cap far more often than BreadthKV's. With the paper's formats and no centering, ThinKV is 22.1\pp{} below BreadthKV over the 14 settings we ran and collapses on R1-Qwen. Part of the gap may come from our implementation choices, most likely the runtime thresholds and K centering; we did not check it against a setting from the ThinKV paper.

\subsection{Runs that reach the length cap}\label{sec:derail}
Most of what eviction loses on R1-Llama and Qwen3 comes from runs that never finish. We call a run \emph{derailed} if it reaches the 32K cap. For Qwen3-8B this cap is below the 38,912 tokens its model card suggests for competition math,\footnote{\url{https://huggingface.co/Qwen/Qwen3-8B}} and FullKV itself reaches it in 10\% of AIME samples. Continuing every capped Qwen3 AIME run to 38,912 tokens under the same compression (Appendix~\ref{app:cap}) finishes 22 of FullKV's 49 capped runs but only 2 of SnapKV-D's 437 at $B=512$, and BreadthKV's lead does not shrink. Derailed runs are almost never correct (at most 10 of 240--480 samples per configuration are both derailed and correct), and on these two models 74--98\% of SnapKV-D's wrong answers are derailed (Table~\ref{tab:derail}; Fig.~\ref{fig:derail}). On R1-Qwen, derailment accounts for fewer of the errors, and compression also hurts runs that finish; on the same problems, finished SnapKV-D runs at $B=512$ score 21.9\% against 42.5\% for FullKV.

\begin{table}[t]
\centering\footnotesize\setlength{\tabcolsep}{2.5pt}
\caption{Derailment on the test AIMEs. acc: pass@1 (\%). cap: samples hitting the 32K cap (\%). c/w: share of wrong answers that are capped.}\label{tab:derail}
\begin{tabular}{llcccccc}
\toprule
& & \multicolumn{3}{c}{@512} & \multicolumn{3}{c}{@1024} \\
\cmidrule(lr){3-5}\cmidrule(lr){6-8}
model & method & acc & cap & c/w & acc & cap & c/w \\
\midrule
\multirow{4}{*}{R1-Qwen} & FullKV & 41.2 & 6.7 & 11\% & & & \\
& SnapKV-D & 20.4 & 22.9 & 29\% & 26.2 & 20.0 & 27\% \\
& 4-bit hyb. & 29.2 & 22.1 & 31\% & 36.2 & 12.5 & 20\% \\
& 2-bit hyb. & 35.4 & 15.0 & 23\% & 35.4 & 11.2 & 17\% \\
\midrule
\multirow{4}{*}{R1-Llama} & FullKV & 41.2 & 6.0 & 10\% & & & \\
& SnapKV-D & 29.0 & 62.7 & 88\% & 34.4 & 48.5 & 74\% \\
& 4-bit hyb. & 41.9 & 38.8 & 67\% & 42.5 & 17.5 & 30\% \\
& 2-bit hyb. & 37.9 & 17.3 & 28\% & 41.5 & 4.4 & 7\% \\
\midrule
\multirow{4}{*}{Qwen3} & FullKV & 69.8 & 10.2 & 32\% & & & \\
& SnapKV-D & 9.4 & 91.0 & 98\% & 18.3 & 78.3 & 94\% \\
& 4-bit hyb. & 28.8 & 65.4 & 90\% & 57.3 & 37.7 & 86\% \\
& 2-bit hyb. & 51.7 & 40.4 & 82\% & 69.0 & 14.8 & 47\% \\
\bottomrule
\end{tabular}
\end{table}

A derailed run rarely loops word for word. Appendix~\ref{app:example} shows one from Qwen3-8B with SnapKV-D at $B=512$, on an AIME24 problem that FullKV and both hybrids solve in all 8 samples. The run finds the right conditional probability within the first tenth of its output and then checks it again and again, in slightly different words each time, until it is cut off at 32K tokens.

Our working hypothesis, which we have not tested, is that eviction deletes the record of a check along with the result. A long CoT keeps returning to its intermediate results; once a result and its earlier check have gone unattended long enough to be evicted, the model derives the result again, and that derivation is evicted in turn. A low-bit cache keeps several times as many tokens, so the earlier check stays visible, if noisy. The frequency of ``wait'' is consistent with this. On Qwen3 AIME24 at $B=512$, SnapKV-D runs contain 6.4 ``wait''s per thousand tokens, against 5.0 for the 2-bit hybrid and 3.7 for FullKV. A direct test would track whether the tokens of an earlier check are still cached when a new check starts.

At equal bytes, fewer bits per token do mean fewer derailed runs. Going from 4 to 3 to 2 bits never raises the capped share, and 2 bits is below 4 bits in all six model--budget pairs. Fewer derailed runs do not always mean higher accuracy, because fewer bits also add noise to every retained token. On R1-Llama at $B=512$, 4 bits derails about twice as often as 2 bits and still scores higher.

\subsection{The floor of breadth: sub-2-bit and rotated codecs}\label{sec:floor}
To test whether breadth still helps below 2 bits, we use TurboQuant-style codecs (\emph{LM-c}: random-sign Hadamard rotation, per-token fp16 norm, Gaussian Lloyd--Max levels, K's channel mean removed per 64-token group). LM-c2 costs 2.25 bits per element (KIVI-2 costs 2.5), and LM-c1 costs 1.25. Offline, LM-c makes 1 bit usable, with attention errors of 0.27--0.40 against 2.2--5.1 for uniform 1-bit quantization, and LM-c2 matches KIVI-2 with 10\% fewer bytes.

End to end, none of them helps, and with these codecs the benefit of breadth ends at about 2 bits. LM-c1 collapses, with 0--12\% calibration accuracy and 73--100\% of samples capped (we stopped these runs early once this was clear). Over the 18 test settings, LM-c2 is 2.6\pp{} below the KIVI-2 hybrid at equal bytes [$-3.8$, $-1.4$], writes 11\% longer outputs and derails more in all six AIME configurations, although its offline error is the same. Whatever makes the difference, the averaged single-step error we measure does not see it. Calibration would have rejected LM-c2 only for Qwen3.

\subsection{Can derailed runs be rescued after the fact?}\label{sec:rescue}
Since derailed runs fail by never concluding, we tested on the calibration set whether forcing a conclusion helps; it mostly does not. For each capped run of the 2-bit hybrid, we cut the run at the loop onset found by an $n$-gram repetition detector (24-grams, $\ge$60\% repeated within a 1K window); if the detector finds no onset, we drop the last 4K tokens instead. We then append \texttt{</think>} and let the model answer under the same compressed cache (Table~\ref{tab:rescue}). This recovers 15\% of capped runs on Qwen3 and almost none on the R1 models, below the 30\% we had set in advance as the bar for running it on the test sets. Part of the problem is the detector, which almost never fires on finished runs but finds an onset in only 6 of 115 and 8 of 51 capped Qwen3 runs, because most derailed runs repeat reasoning steps and not long exact spans, as in Appendix~\ref{app:example}.

\begin{table}[t]
\centering\footnotesize\setlength{\tabcolsep}{3pt}
\caption{Forced conclusion of capped runs (calibration set, 2-bit hybrid): capped runs, capped runs that become correct (recovered), and calibration accuracy (\%) before $\to$ after.}\label{tab:rescue}
\begin{tabular}{lrrrc}
\toprule
model & $B$ & capped & recovered & accuracy \\
\midrule
R1-Qwen-7B & 512 & 37 & 0 & 28.3 $\to$ 28.3 \\
 & 1024 & 22 & 0 & 33.8 $\to$ 33.8 \\
R1-Llama-8B & 512 & 54 & 1 & 33.8 $\to$ 34.2 \\
 & 1024 & 12 & 0 & 40.0 $\to$ 40.0 \\
Qwen3-8B & 512 & 115 & 17 (15\%) & 43.3 $\to$ 50.4 \\
 & 1024 & 51 & 8 (16\%) & 62.9 $\to$ 66.2 \\
\bottomrule
\end{tabular}
\end{table}

\subsection{Efficiency}\label{sec:eff}
To see whether low-bit storage slows decoding, we implement a Triton split-K decode kernel that reads packed low-bit KV directly. Codes are stored in a planar layout and dot products run in fp16 on tensor cores. It uses the layout of Appendix~\ref{app:pact} (one V scale/zero pair per token instead of two), with a maximum relative error of 0.7\% against a PyTorch reference. On an H200 (batch 64, R1-Qwen-7B heads, one layer, one decode step) at 32K context, fp16 FlashAttention takes 1.63\,ms and int8/int4/int2 take 2.35/1.70/2.21\,ms; at 16K, the times are 0.75 and 1.20/0.87/1.13\,ms. At equal context, int4 is close to fp16 at 27\% of the memory, but int2 is 36\% slower than fp16 at 32K; the kernel is compute-bound (7 query heads per KV head are padded to 16 MMA rows), and 2-bit unpacking adds work. At the operating points, the time follows the number of cached tokens, not their bytes. On an A800 with Qwen3-8B heads, SnapKV-D's fp16 attention over 768 tokens takes 0.16\,ms, the 2-bit kernel over 2,560 tokens (about BreadthKV's cache at $B=512$) 0.59\,ms, and fp16 attention over 16K tokens 3.6\,ms (Appendix~\ref{app:kernel}). BreadthKV holds slightly less KV memory than SnapKV-D and needs 24\% fewer decode steps, but with our kernel its attention takes 3.7--4.4$\times$ as long per step. We did not measure serving throughput.

\section{Conclusion}
For decode-time compression of long CoT, how a byte budget is split between tokens and precision mattered more in our experiments than which tokens an eviction method keeps. Spending the bytes on more low-bit tokens, with the bit-width set by a short end-to-end calibration, beat eviction in 17 of 18 settings and was within noise of a joint rate--distortion allocator that uses 27\% more memory-time. The offline attention error we measured and per-head probes did not identify the bit-width, so we would not set it without an end-to-end run. The open problem is derailment on long traces; on Qwen3-8B, 40\% of 2-bit runs at our tightest budget still reach the cap, and forcing a conclusion recovers few of them. If our account of derailment is right, a cache that keeps even a coarse record of what the model has already checked should help more than a better token scorer.

\section*{Limitations}
\textbf{Calibration.} The calibration set is small (60 problems), so on the R1-Distill models its choice is within noise of the alternatives, and it was validated at two budgets only. At these budgets, always using 2 bits is statistically on par with calibration. The benefit of calibration is that it avoids the model-specific failures of other fixed choices; whether it is needed at larger budgets has not been tested on held-out data.

\textbf{Statistical power.} AIME has 30 problems per year, so single-setting confidence intervals are wide. Cross-setting claims rest on pooled tests. Each main configuration uses one sampling seed. Two more seeds on one setting per model (Appendix~\ref{app:setup}) keep BreadthKV ahead of SnapKV-D, but on the R1-Distill models the calibrated width is not the best one in every seed.

\textbf{Models, tasks and languages.} We study three open 7--8B reasoning models (two R1-distilled, one Qwen3) on English mathematics and science benchmarks; GPQA was run on one model only. We did not evaluate code generation, long-input tasks, other languages, larger models or other reasoning-model families, so the breadth--depth trade-off and the calibrated bit-widths may differ there.

\textbf{Mechanism.} The derailment analysis is correlational, and the mechanism in \S\ref{sec:derail} is a hypothesis we have not tested. The rescue experiment used a literal-repetition detector, ran on calibration data only and intervened late. Derailment is defined by a length cap, which we raised to the 38,912 tokens suggested for Qwen3-8B only on Qwen3 AIME runs (Appendix~\ref{app:cap}).

\textbf{Reproduced baselines.} RDKV's decode mode has no total budget, so its memory is matched only approximately, and we modified the zero point of its quantizer (we also report as-released results). ThinKV has no public code; our re-implementation fixes details the paper leaves open, calibrates thresholds on runtime sparsity and adds K-channel centering, whose per-segment means our byte count only partly charges (we report a bound). Results from the authors' own implementations may differ.

\textbf{Scope of the comparisons.} The sub-2-bit study covers one codec family; vector or learned codebooks might move the floor. The low-rank baseline uses a frozen basis. We did not run every recent reasoning-specific method; R-KV and VaSE represent reasoning-aware eviction.

\textbf{Simulated quantization.} Quantization is simulated in the engine (quantize, then dequantize) with exact byte accounting. We do not measure end-to-end memory or serving throughput, and the kernel study (\S\ref{sec:eff}) uses a closely related but not identical layout and is compute-bound.

\textbf{Test contamination.} Some benchmarks, MATH500 in particular, may appear in the models' training data. All comparisons are between compression methods on the same model, so contamination should affect absolute accuracy more than the differences we report, but we cannot rule out interactions.

\section*{Ethical Considerations}
This work studies inference efficiency and poses low direct risk. Compression can preserve final answers while degrading the reasoning that supports them~\citep{accevidence}, so compressed reasoning models should be evaluated beyond final-answer accuracy before deployment. We use publicly released models and benchmarks for evaluation only, consistent with their intended research use. The DeepSeek-R1-Distill models are MIT-licensed (the Llama-based one is also subject to the Llama~3.1 license), Qwen3-8B is under Apache~2.0, GPQA under CC~BY~4.0 and MATH under MIT, and the AIME problems are published by the MAA; we use public copies on Hugging Face. The RDKV and VaSE code is MIT-licensed. We do not redistribute the benchmark data, which contains no personal data. Our generation runs took about 404 GPU-hours (Appendix~\ref{app:setup}).

\bibliography{custom}

\appendix

\section{Experimental Details}\label{app:setup}
\textbf{Prompts.} Each problem is sent as a single user turn through the model's chat template (Qwen3 with thinking enabled), followed by ``Please reason step by step, and put your final answer within \texttt{\textbackslash boxed\{\}}.'' For GPQA the suffix asks for the letter of the correct option (A--D) within \texttt{\textbackslash boxed\{\}}; each question lists its four options as lettered choices.

\textbf{Grading.} We take the last \texttt{\textbackslash boxed\{\}} after \texttt{</think>} (or anywhere in the output if the answer region has none). Math answers are compared with math-verify (0.9.0), falling back to exact string match after removing spaces; GPQA answers are compared by option letter. Runs that hit the length cap without an answer count as wrong. Across the main AIME runs, 76 capped samples were scored correct from a \texttt{\textbackslash boxed\{\}} inside unfinished reasoning, which slightly favors methods that reach the cap more often.

\textbf{Data.} MATH500 (500 problems; the subset of MATH~\citep{mathds} selected by \citealp{lightman}; \texttt{HuggingFaceH4/MATH-500}), AIME 2024 and 2025 (30 each) and GPQA-Diamond (198; \citealp{gpqa}) are loaded from public Hugging Face copies. The calibration set is the 60 AIME 2022--2023 problems of the AIMO validation set (\texttt{AI-MO/aimo-validation-aime}), disjoint from all test sets. AIME problems are published by the Mathematical Association of America.

\textbf{Models and prompts.} R1-Distill-Qwen-7B, R1-Distill-Llama-8B and Qwen3-8B have 7.6B, 8.0B and 8.2B parameters (bf16 weights in Table~\ref{tab:opmem}). With the R1-Qwen tokenizer, prompts average 91 tokens on MATH500, 195 on AIME25 and 247 on GPQA (at most 2,799).

\textbf{Engine and software.} Models run in bf16 in our own decoding engine (FlashAttention-2 decode, CUDA graphs, per-head variable-length caches) with PyTorch 2.8.0, Transformers 4.56.2, FlashAttention 2.8.3 and Triton 3.4.0. Generation runs used H200 and A800 GPUs and took about 404 GPU-hours in total, excluding offline analyses.

\textbf{Run integrity.} Each run stores its configuration, code commit and per-sample generations, and every number in the paper is recomputed from these files by scripts. A scan for duplicate or truncated sample records, which arose once when two workers resumed the same run, found three affected configurations. Duplicates were removed by sample id, and the one affected result was discarded and rerun. The engine is unit-tested against HF (\S\ref{sec:exp}), and our RDKV allocation against the released code (\S\ref{sec:rdkv}).

\textbf{Eviction details.} SnapKV-D scores decode tokens by the attention of the last 8 queries, max-pooled with kernel 7, and always keeps the 8 most recent tokens. All methods compress every 128 decode steps, the interval used in R-KV's released code. Our R-KV score is a port of that code (attention mixed with key redundancy, $\lambda=0.1$). VaSE follows its released VaSE-AttnV script (commit a7a523d): a 64-token recent buffer, eviction every 64 steps, a quarter of the budget for the tokens whose value vectors have the largest range, and the rest sampled without replacement in proportion to the attention of the last 64 queries, average-pooled with kernel 5. Its four sink tokens are covered by keeping the prompt. Pooled over the 18 test settings, VaSE writes outputs 1.01$\times$ as long as SnapKV-D's and is 18.1\pp{} below BreadthKV [$-19.9$, $-16.2$] while holding 1.51$\times$ its KV memory-time.

\textbf{Where settings come from.} The 128-token fp16 window and 64-token quantization groups are fixed for all hybrids. PACT-B's width table and $\gamma$ were chosen on the dev set (Appendix~\ref{app:pact}); RDKV's block budget $\beta$ and ThinKV's thresholds were set on the calibration set (\S\ref{sec:rdkv}, \S\ref{sec:thinkv}).

\textbf{Statistics.} Our primary tests are the pooled comparisons of BreadthKV with SnapKV-D and of calibration with fixed widths; per-setting and per-model differences are descriptive. Length and memory-time ratios are computed per setting and averaged arithmetically over settings; geometric means differ by at most 0.02.

\textbf{Seed variance.} The main runs use sampling seed 0. We reran SnapKV-D and the 4-, 3- and 2-bit hybrids for every model on AIME25 at $B=512$ with seeds 1 and 2 (Table~\ref{tab:seeds}). The standard deviation of accuracy over the three seeds is at most 2.8 points, and BreadthKV's gain over SnapKV-D is positive in every seed (+6.7 to +36.2 points). Qwen3 prefers 2 bits in every seed. On R1-Llama the best width differs in each seed, and the means of the three widths are within 1.1 points. On R1-Qwen, 2 bits is best in every seed and 2.3 points above BreadthKV's 3 bits on average (2.1 with seed 0).

\begin{table}[t]
\centering\footnotesize\setlength{\tabcolsep}{4pt}
\caption{AIME25 at $B=512$ with sampling seeds 0, 1 and 2 (seed 0 is the main run): pass@1 (\%) as mean $\pm$ sd over seeds, the range of BreadthKV's gain over SnapKV-D (pp) across seeds, and the best width in each seed. BreadthKV's width is underlined.}\label{tab:seeds}
\begin{tabular}{lccc}
\toprule
& R1-Qwen & R1-Llama & Qwen3 \\
\midrule
SnapKV-D & 20.0 $\pm$ 1.5 & 24.9 $\pm$ 1.2 & 6.0 $\pm$ 0.2 \\
4-bit & 29.6 $\pm$ 0.7 & 31.2 $\pm$ 1.1 & 21.0 $\pm$ 0.6 \\
3-bit & \underline{31.5 $\pm$ 1.7} & \underline{32.2 $\pm$ 0.6} & 29.6 $\pm$ 0.8 \\
2-bit & 33.8 $\pm$ 1.5 & 31.1 $\pm$ 1.6 & \underline{40.3 $\pm$ 2.8} \\
\midrule
BreadthKV gain & +8.8--12.9 & +6.7--7.9 & +31.3--36.2 \\
best width & 2, 2, 2 & 3, 4, 2 & 2, 2, 2 \\
\bottomrule
\end{tabular}
\end{table}

\section{PACT: Per-Head Online Allocation}\label{app:pact}
PACT was our first attempt to choose the bit-width online, separately for each KV head. It is a negative result, and we report it here for completeness.

\textbf{Format and selection.} PACT uses the hybrids' K format, with V quantized per token in a Hadamard-rotated basis ($b+0.25$ bits); quantizing K per token gives a relative attention error of 1.92 at layer 0 at 2 bits, even after rotation. At each compression step, a head keeps its top-$K_b$ tokens by attention from the last 32 queries, stored at $b$ bits, where $K_b$ is the number of tokens the budget can hold at that width.

\textbf{Variants.} PACT-B reads $b$ from a budget-indexed table chosen on the dev set (512$\to$3, 1024$\to$4, 2048$\to$8). PACT-LA tries each $b\in\{16,8,4,3,2\}$ that is not above the head's current width, measures the relative change in the attention output of the last 32 queries (the probe), and picks the $b$ with the smallest change, evaluating each $b$ at the number of tokens it could keep once the history has grown by a factor of $\gamma=4$. A head's precision never increases. Without this look-ahead, the probe picks 8 bits for 88\% of heads and inflates output lengths.

\textbf{Offline motivation.} With per-token-independent quantization on the pilot traces, letting each head choose its own bit-width lowers attention error by 20--42\% relative to the best single width shared by all heads, and a probe on the last 64 queries recovers 90--95\% of this per-head oracle gain.

\textbf{Dev results (R1-Qwen AIME24).} With one width for all heads, the best width rises with the budget, from 2 bits at $B=256$ (26.2\%) to 3 bits at 512 (46.2\%, against 40.0\% at 4 bits), 4 bits at 1024 (54.2\%) and 8 bits at 2048 (54.6\%). At $B=1024$, PACT-LA scores 52.5\%, PACT-LA with $\gamma=2$ 49.6\%, and the probe without look-ahead 46.7\% with 16.2\% of samples capped, all below the fixed 4-bit width.

\textbf{Test results.} PACT-LA and PACT-B differ significantly in 1 of 18 test comparisons. On Qwen3-8B, PACT-B's table (3 or 4 bits) trails the 2-bit hybrid in all six AIME and MATH500 settings ($-2.0$ to $-18.8$\pp{}, 5 significant), and on R1-Qwen AIME25 at $B=2048$ its 8 bits (34.2\%) trail the 4-bit hybrid (37.9\%).

\section{ThinKV Re-implementation Details}\label{app:thinkv}
The paper leaves open the ternary quantizer (we use TWN), the K-means initialization (we run 8 Lloyd iterations), and how a segment's sparsity is aggregated (we take the mean over its steps). On our models, the paper's KDE calibration rarely finds three sparsity modes (at most 1 of 24 traces per layer), so the thresholds fall back to tertiles. Thresholds computed offline on FullKV traces also do not transfer to decoding under compression, because sparsity depends on the number of cached entries. With offline thresholds, 61--100\% of runtime segments fall into a single type and almost none into transitions. We instead set the thresholds at the tertiles of ThinKV's own runtime sparsity on the calibration set, iterating twice.

\section{Full Results}\label{app:full}
Table~\ref{tab:full} lists every held-out test setting, and Table~\ref{tab:opmem} gives the KV memory a sequence holds at the operating point.

\begin{table*}[t]
\centering\footnotesize\setlength{\tabcolsep}{2.1pt}
\caption{Accuracy (\%) on every held-out test setting, with the share of samples hitting the length cap (\%; 32K tokens, MATH500 16K) in parentheses. VaSE: VaSE-AttnV. RDKV: block-wise, fixed quantizer. ThinKV: with K centering; ThinKV-P: the paper's formats, not run on four R1-Qwen settings. FullKV does not depend on $B$.}\label{tab:full}
\begin{tabular}{llrccccccccc}
\toprule
model & bench. & $B$ & FullKV & SnapKV-D & VaSE & 4-bit & 2-bit & calibrated & RDKV & ThinKV & ThinKV-P \\
\midrule
R1-Qwen & AIME25 & 512 & 41.2 (7) & 20.4 (23) & 20.0 (29) & 29.2 (22) & 35.4 (15) & 33.3 (15) & 28.3 (21) & 23.8 (40) & 0.0 (23) \\
R1-Qwen & AIME25 & 1024 & 41.2 (7) & 26.2 (20) & 28.3 (25) & 36.2 (12) & 35.4 (11) & 36.2 (12) & 38.3 (10) & 30.8 (26) & 0.0 (33) \\
R1-Qwen & MATH500 & 512 & 93.0 (4) & 82.4 (12) & 80.4 (13) & 89.0 (8) & 92.0 (5) & 91.0 (4) & 88.8 (5) & 87.2 (10) & -- \\
R1-Qwen & MATH500 & 1024 & 93.0 (4) & 86.8 (8) & 88.0 (9) & 92.6 (4) & 91.2 (4) & 92.6 (4) & 91.4 (5) & 91.0 (5) & -- \\
R1-Qwen & GPQA & 512 & 36.4 (5) & 23.7 (6) & 24.7 (15) & 29.7 (6) & 34.0 (3) & 31.1 (6) & 30.8 (5) & 27.1 (15) & -- \\
R1-Qwen & GPQA & 1024 & 36.4 (5) & 29.4 (6) & 30.3 (7) & 36.2 (3) & 36.9 (1) & 36.2 (3) & 35.7 (2) & 34.7 (4) & -- \\
\midrule
R1-Llama & AIME24 & 512 & 51.2 (6) & 31.7 (60) & 27.5 (65) & 53.3 (32) & 45.8 (15) & 52.5 (27) & 45.4 (19) & 27.5 (74) & 26.2 (65) \\
R1-Llama & AIME24 & 1024 & 51.2 (6) & 41.7 (46) & 39.6 (48) & 50.4 (15) & 50.0 (3) & 50.0 (3) & 48.3 (4) & 45.0 (40) & 37.5 (28) \\
R1-Llama & AIME25 & 512 & 31.2 (6) & 26.2 (65) & 21.7 (68) & 30.4 (45) & 30.0 (19) & 32.9 (32) & 31.7 (25) & 21.2 (72) & 21.7 (68) \\
R1-Llama & AIME25 & 1024 & 31.2 (6) & 27.1 (51) & 28.3 (54) & 34.6 (20) & 32.9 (5) & 32.9 (5) & 35.8 (5) & 29.2 (45) & 28.7 (38) \\
R1-Llama & MATH500 & 512 & 89.4 (5) & 82.8 (11) & 80.0 (15) & 88.2 (8) & 88.0 (6) & 89.6 (6) & 86.4 (7) & 82.8 (15) & 84.0 (12) \\
R1-Llama & MATH500 & 1024 & 89.4 (5) & 87.0 (8) & 86.4 (10) & 89.6 (5) & 86.2 (6) & 86.2 (6) & 89.4 (5) & 88.8 (6) & 86.6 (7) \\
\midrule
Qwen3 & AIME24 & 512 & 74.2 (6) & 12.9 (88) & 11.2 (89) & 35.8 (58) & 62.1 (30) & 62.1 (30) & 61.3 (33) & 21.7 (78) & 16.2 (87) \\
Qwen3 & AIME24 & 1024 & 74.2 (6) & 25.0 (74) & 26.7 (72) & 64.2 (30) & 75.0 (10) & 75.0 (10) & 72.9 (16) & 43.8 (50) & 32.5 (70) \\
Qwen3 & AIME25 & 512 & 65.4 (14) & 5.8 (94) & 6.2 (95) & 21.7 (73) & 41.2 (51) & 41.2 (51) & 47.9 (50) & 14.6 (86) & 11.7 (90) \\
Qwen3 & AIME25 & 1024 & 65.4 (14) & 11.7 (82) & 13.3 (83) & 50.4 (45) & 62.9 (19) & 62.9 (19) & 60.8 (25) & 27.5 (65) & 20.0 (78) \\
Qwen3 & MATH500 & 512 & 94.8 (5) & 73.0 (28) & 71.0 (31) & 84.8 (15) & 90.4 (9) & 90.4 (9) & 86.4 (14) & 82.6 (17) & 77.2 (23) \\
Qwen3 & MATH500 & 1024 & 94.8 (5) & 81.8 (17) & 81.0 (19) & 92.2 (7) & 93.4 (5) & 93.4 (5) & 91.6 (7) & 89.6 (10) & 87.2 (13) \\
\bottomrule
\end{tabular}

\end{table*}

\section{Longer Length Cap for Qwen3-8B}\label{app:cap}
Qwen3-8B's model card suggests up to 38,912 output tokens for competition math, above our 32K cap. For each Qwen3 AIME configuration in Table~\ref{tab:cap}, we replayed every sample that reached the cap through the same engine and compression method, feeding its 32K stored tokens one decode step at a time so that eviction and quantization act at the same steps as in the original run, and then sampled on to 38,912 tokens with a new seed. Samples that finished under 32K are kept as they are, and extended samples are graded on their full output. Outputs are stored as text, so the replayed prefix is re-tokenized and can differ from the original tokens at a few boundaries.

Under the longer cap, 22 of FullKV's 49 capped samples finish and its capped share falls from 10.2\% to 5.6\%. The longer cap also ends 25 of BreadthKV's 194 capped samples at $B=512$ (40.4\% to 35.2\% capped) and 27 of 71 at $B=1024$, but only 2 of SnapKV-D's 437 at $B=512$ and 10 of 376 at $B=1024$. Accuracy moves by at most 1.3 points for any configuration. BreadthKV's lead over SnapKV-D is +42.3 points [32.7, 51.9] with the 32K cap and +43.5 [33.8, 53.8] with the longer one at $B=512$, and +50.6 with both caps at $B=1024$ (paired bootstrap over problems within each year). Almost none of the runs that eviction derails at 32K were close to finishing.

\begin{table}[t]
\centering\small\setlength{\tabcolsep}{4pt}
\caption{Qwen3-8B on AIME24 and AIME25 (480 samples per row) with the 32K cap and with capped runs continued to 38,912 tokens: pass@1 (\%) and samples hitting the cap (\%). BreadthKV uses 2 bits at both budgets.}\label{tab:cap}
\begin{tabular}{lrcccc}
\toprule
& & \multicolumn{2}{c}{pass@1} & \multicolumn{2}{c}{capped} \\
\cmidrule(lr){3-4}\cmidrule(lr){5-6}
method & $B$ & 32K & 38.9K & 32K & 38.9K \\
\midrule
FullKV & -- & 69.8 & 70.6 & 10.2 & 5.6 \\
SnapKV-D & 512 & 9.4 & 9.4 & 91.0 & 90.6 \\
BreadthKV & 512 & 51.7 & 52.9 & 40.4 & 35.2 \\
SnapKV-D & 1024 & 18.3 & 19.6 & 78.3 & 76.2 \\
BreadthKV & 1024 & 69.0 & 70.2 & 14.8 & 9.2 \\
\bottomrule
\end{tabular}
\end{table}

\section{Kernel Timing at the Operating Points}\label{app:kernel}
Table~\ref{tab:kernel} gives the time of one layer's attention for one decode step at batch 64 on an A800 (median of 7 repetitions of 50 calls), at cache sizes near the mean cache of each method on the test AIMEs, prompt included. The kernel holds every cached token at the given width, whereas BreadthKV keeps the prompt and the 128-token window in fp16, so a real BreadthKV step mixes the two. At equal token count, the 4-bit kernel takes 0.66--0.91$\times$ the time of fp16 FlashAttention from 1.3K tokens on (Qwen3 heads) or 1.7K (R1-Qwen heads), the 8- and 2-bit kernels take 0.84--1.21$\times$ as long, and all three are slower on smaller caches. The kernel is limited by compute, not by memory reads, so its time grows with the number of cached tokens. At the same bytes, the 2-bit hybrid's attention takes 3.2--4.4$\times$ as long as SnapKV-D's and the 4-bit hybrid's 2.0--2.2$\times$, while both stay far below fp16 attention over a long trace.

\begin{table*}[t]
\centering\footnotesize
\caption{Decode-attention time (ms) of one layer for one decode step at batch 64 on an A800, by number of cached tokens and storage width (fp16: FlashAttention; int8/4/2: our Triton kernel), with the attention heads of R1-Qwen-7B (4 KV, 28 query) and Qwen3-8B (8 KV, 32 query). Token counts are near the mean cache of each method on the test AIMEs, prompt included; the kernel stores all of them at the given width. The Qwen3 run at 32K ran out of memory while quantizing the test cache.}\label{tab:kernel}
\begin{tabular}{rlcccccccc}
\toprule
& & \multicolumn{4}{c}{R1-Qwen-7B heads} & \multicolumn{4}{c}{Qwen3-8B heads} \\
\cmidrule(lr){3-6}\cmidrule(lr){7-10}
tokens & near the cache of & fp16 & int8 & int4 & int2 & fp16 & int8 & int4 & int2 \\
\midrule
768 & SnapKV-D, $B=512$ & 0.10 & 0.19 & 0.19 & 0.19 & 0.16 & 0.23 & 0.19 & 0.21 \\
1,280 & SnapKV-D, $B=1024$ & 0.17 & 0.19 & 0.19 & 0.19 & 0.29 & 0.35 & 0.26 & 0.32 \\
1,664 & 4-bit hybrid, $B=512$ & 0.21 & 0.24 & 0.20 & 0.23 & 0.38 & 0.43 & 0.32 & 0.39 \\
2,560 & 2-bit hybrid, $B=512$ & 0.33 & 0.35 & 0.25 & 0.31 & 0.58 & 0.63 & 0.47 & 0.59 \\
3,456 & 4-bit hybrid, $B=1024$ & 0.44 & 0.45 & 0.34 & 0.41 & 0.77 & 0.83 & 0.63 & 0.78 \\
5,888 & 2-bit hybrid, $B=1024$ & 0.75 & 0.72 & 0.53 & 0.66 & 1.30 & 1.37 & 1.01 & 1.26 \\
16,384 & FullKV, long trace & 2.05 & 1.88 & 1.38 & 1.72 & 3.57 & 3.70 & 2.71 & 3.40 \\
32,768 & FullKV, long trace & 4.10 & 3.69 & 2.71 & 3.43 & -- & -- & -- & -- \\
\bottomrule
\end{tabular}
\end{table*}

\section{A Derailed Run}\label{app:example}
Qwen3-8B with SnapKV-D at $B=512$, on the AIME24 lottery problem (answer 116). FullKV and the 4- and 2-bit hybrids solve this problem in all 8 samples, using 6--16K tokens; SnapKV-D reaches the cap in 5 of 8. In the sample below, the model first states the conditional probability $1/115$ about 8\% of the way into its output. The sentence ``the conditional probability is 1 / 115'' alone then occurs 23 times before the run is cut off at 32{,}768 tokens without \texttt{</think>}. The excerpt starts 92\% of the way into the output and is verbatim, with [\ldots] marking omissions.

\begin{quote}\small
Therefore, the conditional probability is 1 / 115. However, the problem says that the answer is m/n where m and n are coprime. 1 and 115 are coprime, so m + n is 1 + 115 = 116. But I need to check if this is correct.

Wait, but maybe I made a mistake here. [\ldots] So yes, that seems correct.

But wait, maybe there's a mistake in calculating the number of ways she can win a prize. Let me check again. [\ldots] Which is 6*15 + 4*6 + 1*1 = 90 + 24 + 1 = 115. Yes, that's correct. So the number is 115. Therefore, the conditional probability is 1/115. So m + n is 1 + 115 = 116.

But wait, the problem says ``the probability of her winning the grand prize given that she won a prize''. So that is 1/115? But I have a feeling that maybe I made a mistake here. Let me check with another approach. [\ldots]

Wait, but the problem says ``the probability of her winning the grand prize given that she won a prize''. So that's 1/115. But I'm wondering why the answer is not something else. Maybe I made a mistake in calculating the number of ways she can win a prize? [\ldots]
\end{quote}
The output ends mid-sentence, while the model is computing the same count of 115 once more.

\begin{table*}[t]
\centering\footnotesize\setlength{\tabcolsep}{4pt}
\caption{Mean peak KV cache per sequence on the test AIMEs (GB; all layers, prompt included) and bf16 model weights (GB). BKV: BreadthKV, which holds 5--9\% less peak memory than SnapKV-D at the same $B$.}\label{tab:opmem}
\begin{tabular}{lcccccc}
\toprule
& & & \multicolumn{2}{c}{$B=512$} & \multicolumn{2}{c}{$B=1024$} \\
\cmidrule(lr){4-5}\cmidrule(lr){6-7}
model & weights & FullKV & SnapKV-D & BKV & SnapKV-D & BKV \\
\midrule
R1-Qwen & 15.2 & 0.81 & 0.048 & 0.044 & 0.077 & 0.073 \\
R1-Llama & 16.1 & 1.84 & 0.104 & 0.096 & 0.171 & 0.155 \\
Qwen3 & 16.4 & 2.53 & 0.118 & 0.108 & 0.194 & 0.184 \\
\bottomrule
\end{tabular}
\end{table*}

\end{document}